%% file: eccv2022submission.tex
\documentclass[runningheads]{llncs}
\usepackage{graphicx}
\usepackage{booktabs}
\usepackage{shadowtext}
\usepackage{amssymb}
\usepackage{comment}
\usepackage{amsmath,amssymb} % define this before the line numbering.
\usepackage{sidecap}
\usepackage{color}
\usepackage{amsmath}
\usepackage{caption} 
\usepackage{algorithm,algorithmic}
\usepackage{sidecap, caption}
\usepackage[colorlinks,citecolor=red,urlcolor=blue,bookmarks=false,hypertexnames=true]{hyperref}
\usepackage[accsupp]{axessibility}  % Improves PDF readability for those with disabilities.

\begin{document}
% \linenumbers
\pagestyle{headings}
\mainmatter
\def\ECCVSubNumber{3797}  % Insert your submission number here

\shadowrgb{0.8, 0.8, .8}
\shadowoffset{3pt}
% \title{Synthesizing Objects from Shadows} % Replace with your title

\title{Shadows Shed Light on 3D Objects}

\begin{comment}
\titlerunning{ECCV-22 submission ID \ECCVSubNumber} 
\authorrunning{ECCV-22 submission ID \ECCVSubNumber} 
\author{Anonymous ECCV submission}
\institute{Paper ID \ECCVSubNumber}
\end{comment}
%******************

% \begin{comment}
\titlerunning{Shadows Shed Light on 3D Objects}
\author{Ruoshi Liu\inst{1} \and
Sachit Menon\inst{1} \and
Chengzhi Mao\inst{1} \and
Dennis Park\inst{2} \and \\
Simon Stent\inst{2} \and
Carl Vondrick\inst{1}}
\authorrunning{R. Liu et al.}
\institute{Columbia University \and Toyota Research Institute\\
\email{\{ruoshi.liu, sachit.menon, cm3797, cv2428\}@columbia.edu}\\
\email{dennis.park@tri.global, }
\email{sistent@gmail.com}}
% \end{comment}
%******************
\maketitle

\input{def.tex}

\begin{abstract}
3D reconstruction is a fundamental problem in computer vision, and the task is especially challenging when the object to reconstruct is partially or fully occluded. We introduce a method that uses the shadows cast by an unobserved object in order to infer the possible 3D volumes behind the occlusion. 
We create a differentiable image formation model that allows us to jointly infer the 3D shape of an object, its pose, and the position of a light source. Since the approach is end-to-end differentiable, we are able to integrate learned priors of object geometry in order to generate realistic 3D shapes of different object categories. 
%Our approach works when the object itself may not be visible, and there is only a silhouette on the floor created by the object blocking light.
Experiments and visualizations show that the method is able to generate multiple possible solutions that are consistent with the observation of the shadow. Our approach works even when the position of the light source and object pose are both unknown. Our approach is also robust to real-world images where ground-truth shadow mask is unknown.

\end{abstract}

\section{Introduction}

\begin{figure}[b]
    \centering
    \includegraphics[width=0.9\linewidth]{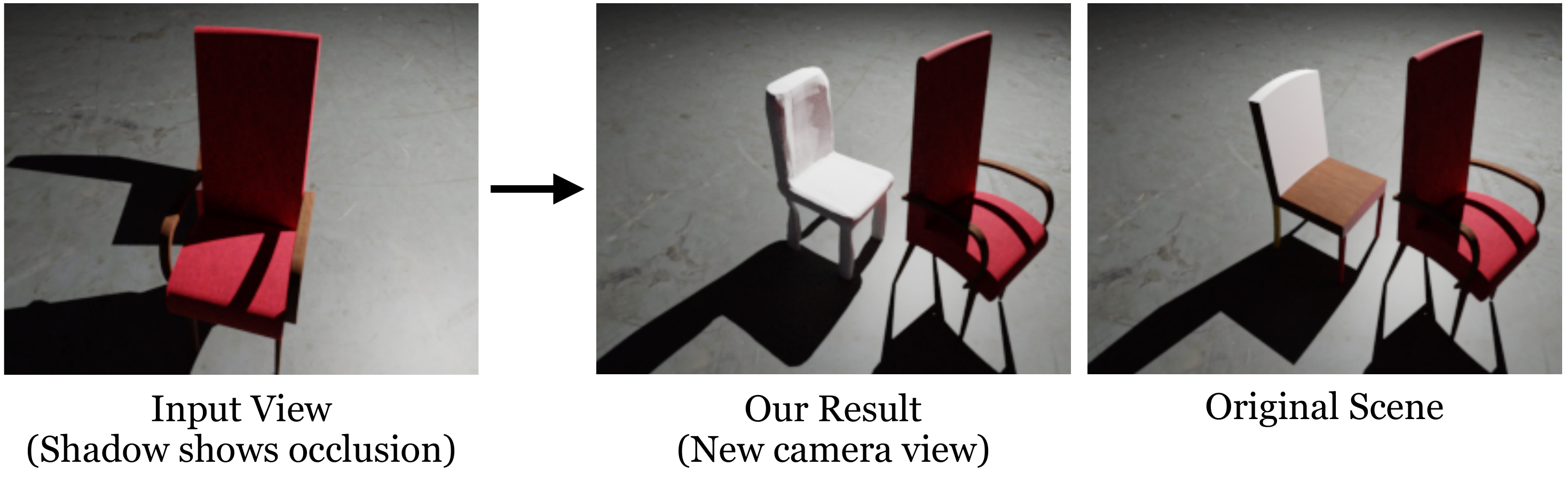}
    \caption{We introduce a method to perform 3D reconstruction from the shadow cast on the floor by occluded objects. In the middle, we visualize our reconstruction of the occluded chair from a new camera view.}
    \label{fig:teaser}
\end{figure}

Reconstructing the 3D shape of objects is a fundamental challenge in computer vision, with a number of applications in robotics, graphics, and data science. The task aims to estimate a 3D model from one or more camera views, and researchers over the last twenty years have developed excellent methods to reconstruct visible objects \cite{horry1997tour,hoiem2005automatic,ye2021shelf,mescheder2019occupancy,mildenhall2020nerf,hartley2003multiple,agarwal2010bundle}. However, objects are often occluded, with the line of sight obstructed either by another object in the scene, or by themselves (self-occlusion).
%When there are no appearance features due to an occlusion, we need to rely on additional context for reconstruction.
Reconstruction from a single image is an under-constrained problem, and occlusions further reduce the number of constraints.
To reconstruct occluded objects,  we need to rely on additional context.

One piece of evidence that people use to uncover occlusions is the shadow cast on the floor by the hidden object. For example, Figure \ref{fig:teaser} shows a scene with an object that has become fully occluded.  Even though no appearance features are visible, the shadow reveals that another object exists behind the chair, and the silhouette constrains the possible 3D shapes of the occluded object. What hidden object caused that shadow?

In this paper, we introduce a framework for reconstructing 3D objects from their shadows. We formulate a generative model of objects and their shadows cast by a light source, which we use to jointly infer the 3D shape and the location of the light source. Our model is fully differentiable, which allows us to use gradient descent to efficiently search for the best shapes that explain the observed shadow. Our approach integrates both learned empirical priors about the geometry of typical objects and the geometry of cameras in order to estimate realistic 3D volumes that are often encountered in the visual world.

%Our approach is based on a generative model of objects and their shadows, which we show we can invert in order to infer possible 3D shapes and possible light sources given the observation of a shadow. Our model is fully differentiable, allowing us to obtain solutions through gradient descent.  3D reconstruction is an inverse problem, and our approach is able to condition on the observed shadow silhouette and use priors about object geometry in order to estimate multiple plausible 3D volumes that are consistent with the shadow. 

%Our entire model is differentiable, allowing our inference algorithm to search for 3D meshes that explain the shadows using gradient descent.  

Since we model the image formation process, we are able to jointly reason over the object geometry and the parameters of the light source. When the light source is unknown, we recover multiple different shapes and multiple different positions of the light source that are consistent with each other. When the light source location is known, our approach can make use of that information to further refine its outputs.
%Moreover, experiments and visualizations show that modeling image formation provides advantages for robustness, allowing the system to generalize to situations outside of the training set, such as new camera views or different types of occlusions.
We validate our approach for a number of different object categories on a new ground truth dataset that we will publicly release.

The primary contribution of this paper is a method to use the shadows in a scene to infer the 3D structure, and the rest of the paper will analyze this technique in detail. Section 2 provides a brief overview of related work for using shadows. Section 3 formulates a generative model for objects and how they cast shadows, which we are able to invert in order to infer shapes from shadows. Section 4 analyzes the capabilities of this approaches with a known and unknown light source. We believe the ability to use shadows to estimate the spatial structure of the scene will have a large impact on computer vision systems' ability to robustly handle occlusions.

\section{Related Work}

We briefly review related work in 3D reconstructions, shadows, and generative models. Our paper combines a model of image formation with generative models.

\textbf{Single-view 3D Reconstruction and 3D Generative Models:}
The task of single-view 3D reconstruction -- given a single image view of a scene or object, generate its underlying 3D model -- has been approached by deep learning methods in recent years. This task is related to unconditional 3D model generation; while unconditional generation creates 3D models a priori, single-view reconstruction can be thought of as generation a posteriori where the condition is the input image view. Given the under-constrained nature of the problem, this is usually done with 3D supervision. Different lines of work address this by generating 3D models in different types of representations \cite{shin_pixels_2018}: specifically, whether they use voxels \cite{brock_generative_2016}, point cloud representations \cite{fan_point_2016}, meshes \cite{groueix2018papier,pontes2018image2mesh}, or the more recently introduced \textit{occupancy networks} \cite{mescheder2019occupancy}. While our general approach is compatible with any of these types of 3D representations, we elect to use occupancy networks in our implementation here as they achieve excellent performance at generating 3D structure. %Occupancy networks enjoy the advantages of X, Y, and Z. 

Occupancy networks~\cite{mescheder2019occupancy} learn a function mapping from 3D position to a scalar describing the probability of occupancy -- that is, whether part of the 3D object occupies that position or not. This function is parameterized by a neural network. The learned occupancy function can be thought of as a binary classifier for each point, and the decision boundary of this classifier then provides a description of the object's surface. To extend this to the generative model setting, a latent vector with a Gaussian prior can be provided to the occupancy function in conjunction with the position vector~\cite{mescheder2019occupancy}. This enables sampling different occupancy functions by sampling different latent vectors.

The cost to obtain 3D ground truth for supervision \cite{h36m_pami} poses a great limitation to the single-view 3D reconstruction. To scale up the applications, another line of work uses multi-view 2D images as supervision \cite{niemeyer2020differentiable,humanMotionKanazawa19,yu2020pixelnerf}, or even only single image as supervision \cite{cmrKanazawa18,liu2019soft,ucmrGoel20,wu2020unsupervised,li2020self,ye2021shelf}. More classically, approaches using Multi-View Stereo (MVS) reconstruct 3D object by combining multiple views \cite{bleyer2011patchmatch,de1999poxels,broadhurst2001probabilistic,galliani2016gipuma,schonberger2016pixelwise,seitz2006comparison,seitz1999photorealistic}.

\textbf{Occlusions and Shadows:}
A major challenge towards 3D reconstruction is the existence of occlusions. From a single image, we do not have access to the full 3D structure of what we are viewing, as some parts are covered up either by other objects, or even by the object we are trying to reconstruct itself (self-occlusion). For example, when we view a chair from behind, the back of the chair may occlude the sides from our viewing angle, making it unclear whether the chair has arms or not. 

Shadows present a naturally-occurring visual signal that can help to clarify this uncertainty. By observing the shadows cast by what we cannot see, we gain insight into the 3D structure of the unseen portion. Previous work has considered the use of shadows towards elucidating structure in a classic vision context. \cite{waltz_understanding_1975} first applied shadows to determine shapes in 3D line drawings. This was extended by \cite{shafer_using_1983}, who determined surface orientations for polyhedra and curved surfaces with shadow geometry. Shadows can also be used more actively to recover 3D shape. \cite{bouguet1993shadows} shows how shadows can help infer the geometry of a shaded region. They cast shadows on an object sitting on a plane, by moving a straight-line occlusion (a stick) around in front of a light source, and propose an efficient method to extract the underlying 3D object shape from the moving contours of the shadow. \cite{savarese_3d_2007} also propose \textit{shadow carving}, a way of using multiple images from the same viewpoint but with different lighting conditions to discover object concavities. They prove that shadows can provide information that guarantees conservative and reasonable shape estimates under some assumptions. Meanwhile, \cite{troccoli_shadow_2004} use shadows as cues to determine parameters for refining 3D textures. Recent work has leveraged deep learning tools to enable detection of shadows from realistic images \cite{wang_instance_2020}, making it possible to extend the use of shadows to realistic settings. Thus far, shadows have not seen much application in determining structure using the tools afforded to vision by the latest deep learning techniques. 

\textbf{Generation Under Constraints:}
Generation under constraints appears throughout the literature in many forms. It falls under the general framework of analysis by synthesis \cite{krull2015learning,yuille2006vision}. Tasks such as super-resolution, image denoising, and image inpainting, begin with an incomplete image and ask for possible reconstructions of the complete image \cite{ongie_deep_2020}. In other words, the goal is to generate realistic images that satisfy the constraint imposed by the given information. Typical approaches consider this as conditional generation, where a function (usually, a neural network) is learned to map from corrupted inputs to the desired outputs \cite{dong_image_2015,kim_accurate_2016,ongie_deep_2020}. More recently, \cite{menon2020pulse} propose using search rather than regression to address these types of tasks in the context of the super-resolution problem. They use pretrained generative models as a map of the natural image manifold, using gradient-based search to find points in the latent space that map to high-resolution images that downscale to the appropriate low-resolution. They avoid generating unrealistic outputs by constraining the search in the latent space to a constant deviation from a spherical Gaussian mean.  While posing the inverse problem as search makes inference more costly, it has benefits in realism and the ability to generate multiple solutions. It further eschews the need to learn the generative process from scratch by leveraging the knowledge captured by the unconditional generator. Recent work by~\cite{sadekar2022shadow} uses differentiable rendering to deform an icosphere to generate targeted shadow art sculptures, with interesting results. Unlike their work, ours focuses on generating a set of \textit{plausible} objects which could explain a given shadow in a real scene. Our more general approach also handles the common scenario in which the light source generating the shadow is unknown.

\section{Method}

We represent the observation of the shadow as a binary image $\s \in \mathbb{R}^{W \times H}$. Our goal is to estimate a set of possible 3D shapes, their poses, and corresponding light sources that are consistent with the shadow $\s$. We approach this problem by defining a generative model for objects and their shadows. We will use this forward model to find the best 3D shape that could have produced the shadow.

\subsection{Explaining Shadows with Generative Models} 

Let $\Omega = G(\z)$ be a generative model for 3D objects, where $\Omega$ parameterizes a 3D volume and $\z \sim \mathcal{N}(0, I)$ is a latent vector with an isotropic prior. When the volume blocks light, it will create a shadow. We write the location of the illumination source  as $\c \in \mathbb{R}^3$ in world coordinates, which radiates light outwards in all directions. The camera will observe the shadow $\hat{s} = \pi(c, \Omega)$, where $\pi$ is a rendering of the shadow cast by the volume $\Omega$ onto the ground plane. 

To reconstruct the 3D objects from their shadow, we formulate the problem as finding a latent vector $\z$, object pose $\phi$, and light source location $\c$ such that the predicted shadow $\hat{\s}$ is consistent with the observed shadow $\s$. We perform inference by solving the optimization problem:
\begin{equation}
%\begin{aligned}
     \min_{\z, \c, \phi} \; \mathcal{L}\left(s, \pi(\c, \Omega) \right) \quad \textrm{where} \quad \Omega =  \mathcal{T}_\phi\left(G(\z)\right)
%\end{aligned}
    \label{eqn:inference}
\end{equation}
The loss function $\mathcal{L}$ compares the candidate shadow $\hat{s} = \pi(\c, \Omega)$ and the observed shadow $\s$, and since silhouettes are binary images, we use a binary cross-entropy loss. We model the object pose with an SE(3) transformation $\mathcal{T}$ parameterized by quaternions $\phi$. In other words, we want to find a latent vector that corresponds to an appropriate 3D model of the object that, in the appropriate pose, casts a shadow matching the observed shadow. 
%Consequently, we can freely choose the location of the camera; we do not need to model the camera extrinsic parameters.
% Since $\z$ is normally distributed, we constrain the norm of $\z$ to be within $\delta \in \mathbb{R}$ distance to the surface of a unit hyper-sphere. 

\begin{figure}[t]
    \centering
    \includegraphics[width=\linewidth]{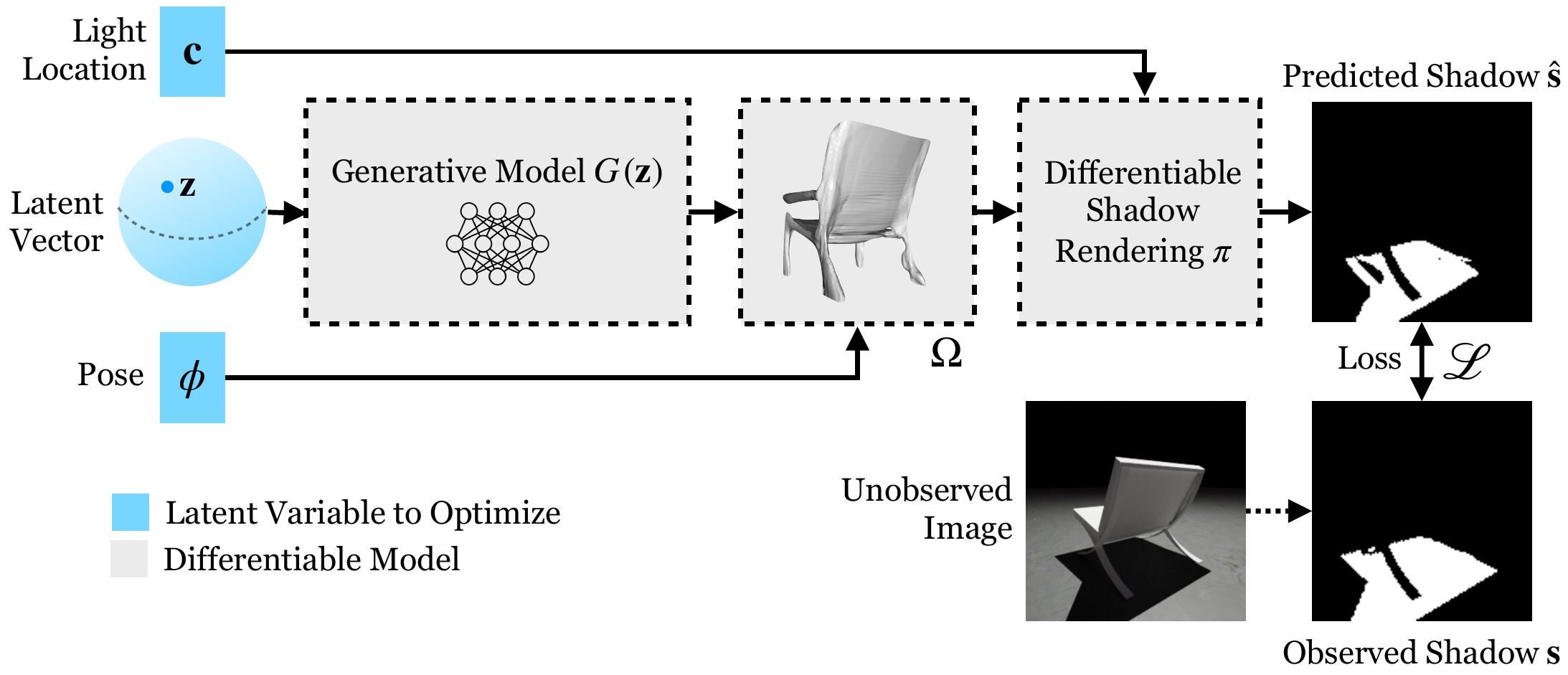}
    \vspace{-2em}
    \caption{\textbf{Overview of our method.} Given an observation of a shadow $\s$, we optimize for an explanation jointly over the location of the light $\c$, the pose of the object $\mathcal{T}_\phi$, and the latent vector of the object 3D shape $\z$. Since every step is differentiable, we are able to solve this optimization problem with gradient descent. By starting the optimization algorithm with different initializations, we are able to recover multiple possible explanations $\Omega$ for the shadow.\vspace{-1em}}
    \label{fig:method}
\end{figure}

Figure \ref{fig:method} illustrates an overview of this setup.
The solution $\z^*$ of the optimization problem will correspond to a volume that is consistent with the observed shadow. We can obtain the resulting shape through $\Omega^* = \mathcal{T}_{\phi^*}\left(G(\z^*)\right)$.
By solving Equation \ref{eqn:inference} multiple times with different initializations, we obtain a set of solutions $\{ \z^ *\}$ yielding multiple possible 3D reconstructions.

\subsection{$G(\z)$: Generative Models of Objects}

%There are many 3D volumes $\Omega$ that will produce a shadow consistent with $\s$, but most of them will be unlikely to naturally occur.

To make the reconstructions realistic, we need to incorporate priors about the geometry of objects typically observed in the visual world.
Rather than searching over the full space of volumes $\Omega$, our approach searches over the latent space $\z$ of a pretrained deep generative model $G(\z)$. Generative models that are trained on large-scale 3D data are able to learn empirical priors about the structure of objects; for example, this can include priors about shape (e.g., automobiles usually have four wheels) and physical stability (e.g., object parts must be supported). By operating over the latent space $\z$, we can use our knowledge of the generative model's prior to constrain our solutions to 3D objects that match the generative model's output distribution.

Our approach is compatible with many choices of 3D representation. In this implementation, we choose to model our 3D volumes with an occupancy field \cite{mescheder2019occupancy}. An occupancy network $\y = f_\Omega(\x)$ is defined as a neural network that estimates the probability $\y \in \mathbb{R}$ that the world coordinates $\x \in \mathbb{R}^3$ contains mass. The generative model $G(\z)$ is trained to produce the parameters $\Omega$ of the occupancy network.

\subsection{$\pi$: Differentiable Rendering of Shadows}

\begin{SCfigure}[1][t]
    \centering
    \includegraphics[width=0.5\linewidth]{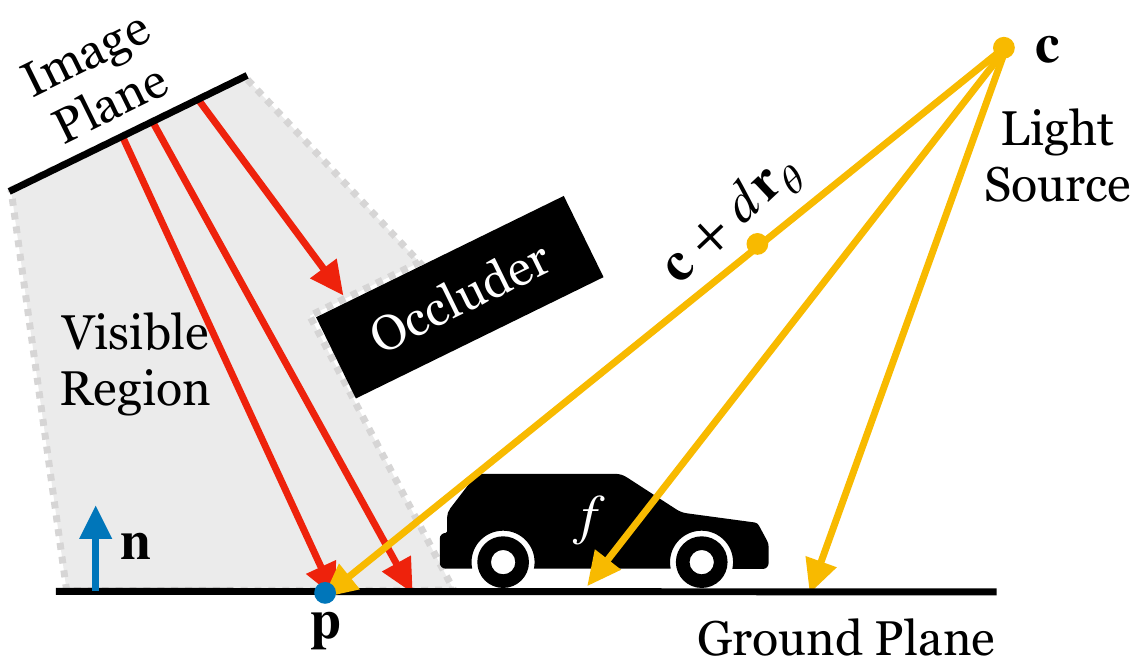}
    \caption{\textbf{Differentiable rendering of shadows.} A point $\p$ on the ground plane will be a shadow if the corresponding ray from the light source intersects with the volume $f$. We calculate whether $\p$ is a shadow by finding the intersecting ray $\r_\theta$, and max pooling $f_\Omega$ along the ray.}
    \label{fig:imageformation}
\end{SCfigure}

To optimize Equation \ref{eqn:inference} with gradient descent, we need to calculate gradients of the shadow rendering $\pi$ and its projection to the camera. This operation can be made differentiable by max-pooling the value of the occupancy network along a light ray originating at the light source. Although integrating occupancy along the ray may be more physically correct to deal with partially transmissive media as in NeRF \cite{mildenhall2020nerf}, since we are primarily concerned with solid, opaque objects and binary shadow masks, we find max-pooling to be a useful simplifying approximation.

Let $\r_\theta \in \mathbb{R}^3$ be a unit vector at an angle $\theta$, and let $\n \in \mathbb{R}^3$ be a vector normal to the ground plane. We need to calculate whether the ray from the light source $\c$ along the direction of $\r_\theta$ will intersect with the ground plane $\n$, or whether it will be blocked by the object $\Omega$. 
%and let $\p_0 \in \mathbb{R}^3$ be a point on the ground plane.
The shadow will be an image  $\pi(\c,\Omega)$ formed on the ground plane, and the intensity on the plane at position $\p$ is given by:
\begin{align}
    \pi(\c,\Omega)[\p] = \max_{d \in \mathbb{R}} \; f_\Omega(\c + d \r_\theta) \quad \textrm{s.t.} \quad  \p = \c - \frac{\c^T\n}{\r_\theta^T \n} \r_\theta 
\end{align}
where we use the notation $\pi(\c,\Omega)[\p]$ to index into $\pi(\c,\Omega)$ at coordinate $\p$. The right-hand constraint between $\p$ and $\r_\theta$ is obtained by calculating the intersection of the light ray with the ground plane.

For the light ray $\r_\theta$ landing at $\p$, the result of $\pi$ is the maximum occupancy value 
$f_\Omega$ along that ray.
Since $\pi(\c,\Omega)$ is an image of the shadow on a plane, it is straightforward to use a homography to transform  $\pi(\c,\Omega)$ into the perspective image $\hat{\s}$ captured by the camera view.

\subsection{Optimization}

Given a shadow $\s$, we optimize $\z$, $\c$, and $\phi$ in Equation \ref{eqn:inference} with gradient descent while holding the generative model $G(\z)$ fixed. We randomly initialize $\z$ by sampling from a multivariate normal distribution, and we randomly sample both a light source location $\c$ and an initial pose $\phi$. We then calculate gradients using back-propagation to minimize the loss between the predicted shadow $\hat{\s}$ and the observed shadow $\s$.

During optimization, we need to enforce that $\z$ resembles a sample from a Gaussian distribution. If this is not satisfied, the inputs to the generative model will no longer match the inputs it has seen during training. This could result in undefined behavior and will not make use of what the generator has learned. We follow the technique from \cite{menon2020pulse}, which made the observation that the density of a high-dimensional Gaussian distribution will condense around the surface of a hyper-sphere (the `Gaussian bubble' effect). By enforcing a hard constraint that $\z$ should be near the hyper-sphere, we can guarantee the optimization will find a solution that is consistent with the generative model prior. 

The objective in Equation \ref{eqn:inference} is non-convex, and there are many local solutions for which gradient descent can become stuck. Motivated by \cite{sgld}, we found that adding linearly decaying Gaussian noise helped the optimization find better solutions. Algorithm \ref{alg:method} summarizes the full procedure.

\begin{algorithm}[t]
\caption{Inference by Inverting the Generative Model}
\label{algorithm: defense}
\begin{algorithmic}[1]
\STATE {\bfseries Input:} Shadow image $\s$, step size $\eta$, number of iterations $K$, and generator $G$.
\STATE {\bfseries Output:} Parameters of a 3D volume $\Omega$
\STATE{\bfseries Inference: }
\STATE{Randomly initialize $\z \sim \mathcal{N}(0, I)$}
\FOR{$k=1,...,K$}
\STATE{$J(\z, \c, \phi) = \mathcal{L}\left(\s, \pi(\c, \mathcal{T}_\phi(G(\z))) \right)$ }
\STATE{$\z \leftarrow \z - \eta \cdot (\nabla_{\z} J(\z, \c, \phi) +  \mathcal{N}(0, \sigma I))$ where $\sigma = \frac{K-1-k}{K}$}
\STATE{$\z \leftarrow \z / ||\z||_2$} 
\STATE{$\c \leftarrow \c - \eta \nabla_{\c} J(\z, \c, \phi)$}
\STATE{$\phi \leftarrow \phi - \eta \nabla_{\phi} J(\z, \c, \phi)$}
\ENDFOR
\STATE{Return parameters of 3D volume $\Omega = \mathcal{T}_\phi(G(\z))$}

% \STATE{$\theta\leftarrow \theta-\eta_2 \sum_{i=1}^m \nabla_\theta[\cL(f_\theta(\x_i),\y_i)+\cL(f_\theta(\x_i),f_\theta(\x_i'))/\lambda]/m$}

\end{algorithmic}
\label{alg:method}
\end{algorithm}

\subsection{Implementation Details}

We will release all code, models, and data. To create $G(\z)$, we use the unconditional 3D generative model from \cite{mescheder2019occupancy}, which is trained to produce an occupancy network with a $128$-dimensional latent vector. The generative model is trained separately on four categories of the ShapeNet dataset, as in~\cite{mescheder2019occupancy}. When the location of the illumination source is unknown, we sample a 3-dimensional coordinate $\c$ from the surface of the northern hemisphere above the ground plane with a fixed radius of 3. When the SE(3) transformation for the object pose is unknown, we sample a 4-dimensional quaternion $\phi$ to parameterize the rotation matrix. A non-zero rotation for ``pitch'' and ``roll'' are physically implausible given a level ground plane and the assumption of upright object, so we constrain them to be zero during optimization. To optimize the full model, we use spherical gradient descent to optimize $\z$ (and optionally $\c$ and $\phi$) for up to 300 steps. We use a step size of 1.0 for known light and pose experiments and 0.01 for unknown light and pose experiments.

To accomplish the differentiable shadow rendering $\pi$, we evenly sample 128 points along each light ray emitted from the illumination source, then evaluate them for occupancy. In the case of occlusion from other objects as well as self-occlusion, we calculate the segmentation mask of all objects in the scene, and disable gradients coming from light rays intersecting with these masks.

% \subsection{Regression: Learning a Inverse Model}

% In order to recover the 3D shape $\mathcal{X}$ from the shadow silhouette $s$, we need to invert the  image formation process for shadows. One way to estimate the inverse model is to approximate it with a neural network $F$ that estimates the latent shape vector $z$ such that the generator $G$ is able to reconstruct the right shape:
% \begin{align}
%     \min_{F,G} \;
%     \mathbb{E}_{(s,\mathcal{X})}\left[
%         \mathcal{L}\left( \hat{\mathcal{X}},  \mathcal{X}\right)
%     \right] \quad \textrm{where} \quad  \hat{\mathcal{X}} = G(F(s))
% \end{align} 
% where $G$ and $F$ are both neural networks, and we assume $X$ is a 3D volumetric representation. Since the camera position and the light locations may be unknown, the encoder $F$ will need to learn to estimate these quantities internally. 

\section{Experimental Results}

The goal of our experiments is to analyze how well our method can estimate 3D shapes that are consistent with an observed shadow. We first introduce a new 3D shadow dataset, then we perform two different quantitative experiments to evaluate the 3D reconstruction performance of our model. We further provide several visualizations and qualitative analysis of our method.

\subsection{Kagemusha: A Dataset of Shadows}
Kagemusha\footnote{Named after Akira Kurosawa's movie, which translates to ``shadow warrior''.} is a dataset of 3D objects and their shadows. The dataset contains four common objects of the ShapeNet dataset \cite{chang2015shapenet}. For each 3D object, we sample a random point light source location from the northern hemisphere with a radius of 3 and a camera location sampled from the northern hemisphere with a radius of 2, to create a scene with shadow. We then compute the segmentation mask and shadow mask of the object. The dataset uses the same train/validation/test split as the original ShapeNet dataset \cite{chang2015shapenet}.

\subsection{Common Experimental Setup}

\textbf{Evaluation Metric:}
We use volumetric IoU to evaluate the accuracy of 3D reconstruction. Volumetric IoU is calculated by dividing the intersection of the two volumes by their union. We uniformly sample 100k points in the bounding volume. We then calculate the occupancy agreement of the points between the candidate 3D volume and the original 3D volume. 

\begin{table}[t]
\centering
\begin{tabular}{p{3.5cm} p{1cm} p{1cm} p{1cm} p{1cm} p{1cm}}
\toprule
Method & Car & Chair & Plane & Sofa & All \\
\midrule
Random & .329 & .203 & .211 & .209 & .238 \\
Nearest Neighbor & .414 & .299 & .349 & .352 & .322 \\
Regression & .611 & .274 & .410 & .524 & .467 \\
Latent Search (Ours) & \textbf{.706} & \textbf{.371} & \textbf{.537} & \textbf{.598} & \textbf{.553} \\
\midrule
\midrule
Im2Mesh (full image) & .737 & .501 & .571 & .680 & .622 \\
\bottomrule
\end{tabular}
\
\caption{Results for 3D reconstruction from the shadows assuming the object pose and the light source position are both known. We report volumetric IoU, and higher is better. The Im2Mesh result shows the performance at 3D reconstruction when the entire image is observable, not just the shadows.}
\label{tab:known}%
\end{table}

\textbf{Baselines:}
To validate our method quantitatively, we selected several baselines for comparison. Since we are analyzing how well generative models can explain shadows in images, we compare against the following approaches.
\textit{(i) Regression:}
An alternative approach to the same problem is to train a regression model to map images of shadows $\s$ to 3D volumes $\Omega$. We modified the occupancy network from \cite{mescheder2019occupancy} to perform this task, which is a widely adopted and highly competitive model for single-view 3D reconstruction. During training and inference, we replace the input RGB image with a shadow image. We also loaded the occupancy decoder pre-trained on ShapeNet\cite{chang2015shapenet} and supervised further training with 3D ground truth. \textit{(ii) Nearest Neighbor:} We experiment with a nearest neighbor approach. For each shadow image $\s$ in the test set, we search in the training set for the object whose shadow $\s'$ minimizes $||\s - \s'||_2$ and use the corresponding 3D object as prediction.
\textit{(iii) Random:} We compute chance by selecting a random 3D object from the training set.
\textit{(iv) Full Image:} For analysis purposes, we also compare against an off-the-shelf 3D reconstruction method~\cite{mescheder2019occupancy} that is able to see the entire image (not just the shadows). Since this approach has more information than our method, we do not expect to outperform it. However, this comparison allows us to quantify the amount of information lost when we only operate with shadows.

\begin{figure}[p]
    \centering
    \includegraphics[width=\linewidth]{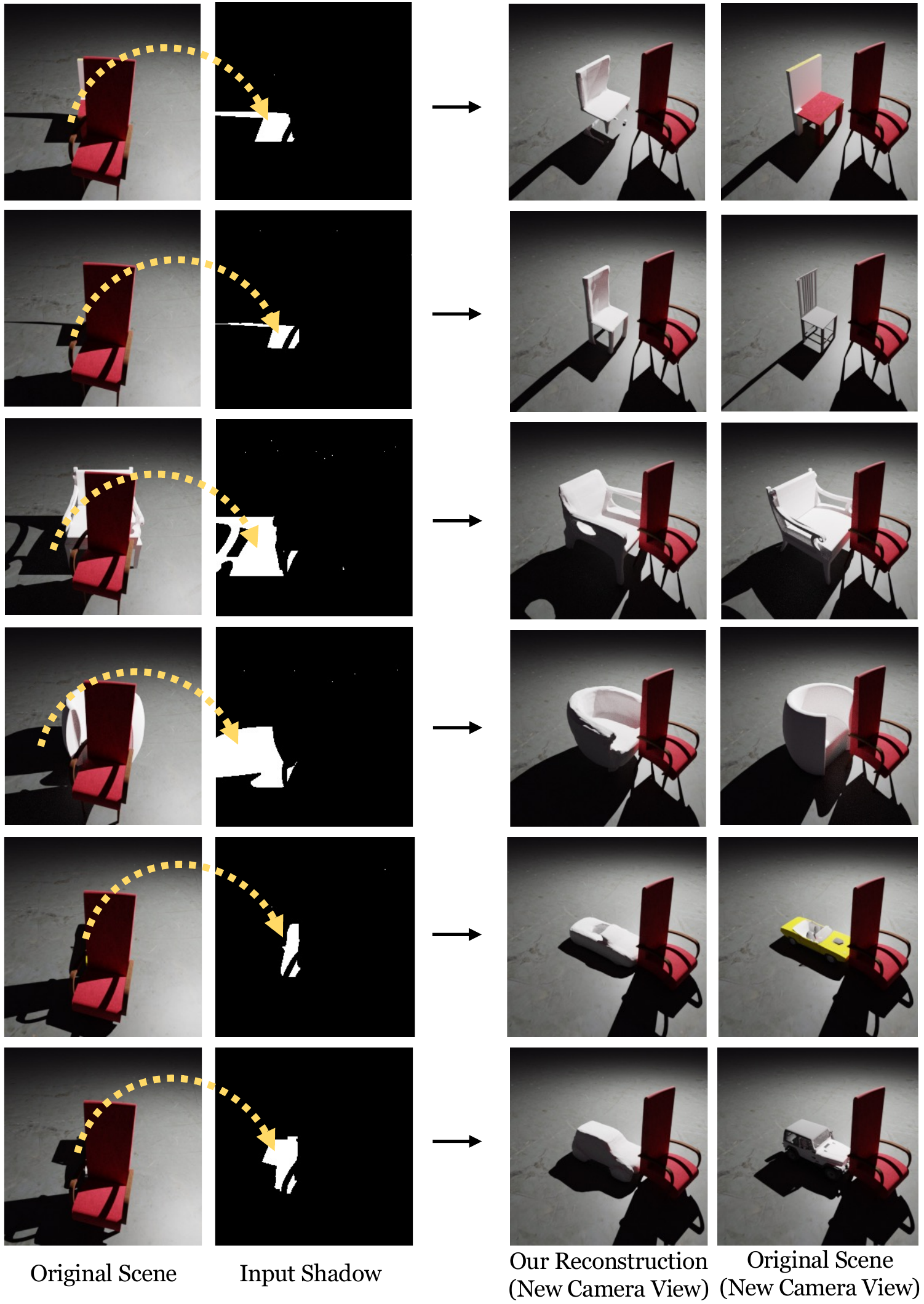}
    \caption{\textbf{3D Reasoning Under Occlusion.} We show several examples of 3D object reconstruction under occlusion. The \textbf{1st} column shows the original scenes including both objects. Shadow masks shown in the \textbf{2nd} column. The \textbf{3rd} and \textbf{4th} column are our reconstruction as seen from another camera view. Note that the red chair in the front is not being reconstructed by our model.} 
    \label{fig:occlusion}
\end{figure}

\subsection{Reconstruction with Known Light and Object Pose}

We first evaluate our method on the task of 3D reconstruction when the light position and pose are known. For each scene, we randomize the location of the light source, and put the objects in their canonical pose. Since the problem is under-constrained, there is not a single unique answer. We consequently run each method eight times to generate diverse predictions, and calculate the average volumetric IoU using the best reconstruction for each example. 

Table \ref{tab:known} compares the performance of our approach versus baselines on this task. Our approach is able to significantly outperform the baselines on this task (by nearly 9 points), showing that it can effectively find 3D object shapes that are consistent with the shadows. Since our approach integrates empirical priors from generative models with the geometry of camera and shadows, it is able to better generalize to the testing set. The regression baseline, for example, does not benefit from these inductive biases, and instead must learn them from data, which our results show is difficult.

When the full image is available, Table \ref{tab:known} shows that established 3D reconstruction methods are able to perform better, which is expected because more information is available. However, when there is an occlusion, the full image will not be available, and we instead must rely on shadows to reconstruct objects.
Figure \ref{fig:occlusion} shows qualitative examples where we were able to reconstruct objects that are occluded other objects. Although there is no appearance information, these results show that shadows allow our model to ``see through'' occlusions in many cases. The examples show that the method is able to reconstruct objects faithfully with diverse shapes and across different categories. We include more examples from all categories in the supplementary materials.

%The main purpose of this evaluation is to experimentally test the effectiveness of the 3D generator and the overall latent search approach. For the convenience of evaluation, we put all the object in its canonical pose and perform differentiable rendering of shadows from a known light source location. We randomly sample latent vector $\z$ as initialization and perform gradient descent as described in the methods. Because our approach can generate diverse 3D objects that are consistent with the input shadow with different initial latent vectors as indicated in \ref{fig:diversity}, for each run, we sample 8 different initialization and picked the output that yields the best result for evaluation.

\subsection{Reconstruction with Unknown Light and Object Pose}

Since our approach is generative and not discriminative, a key advantage is the flexibility to adapt to different constraints and assumptions. In this experiment, we relax our previous assumption that the light source location and the object pose are both known. We evaluate our approach at reconstruction where all three variables (latent vector $\hat{\z}$, light source location $\hat{\c}$, and object pose parameters $\phi$) must be jointly optimized by gradient descent to minimize the shadow reconstruction loss.

\begin{table}[h]
\centering
\begin{tabular}{p{3.5cm} p{1cm} p{1cm} p{1cm} p{1cm} p{1cm}}
\toprule
Method & Car & Chair & Plane & Sofa & All \\
\midrule
Random & .283 & .175 & .177 & .161 & .199  \\
Nearest Neighbor & .346 & .\textbf{233} & .241 & .233 & .264 \\
Regression & .559 & .116 & .218 & .317 & .303 \\
Latent Search (Ours) & \textbf{.618} & .187 & \textbf{.343} & \textbf{.413} & \textbf{.390} \\
\bottomrule
\end{tabular}
\\
\caption{Results for 3D reconstruction from the shadows assuming the object pose and the light source position are \textbf{both unknown}. We report volumetric IoU, and higher is better.\vspace{-1em}}
\label{tab:unknown}%
\end{table}

Table \ref{tab:unknown} shows the performance of our model at reconstructing objects with an unknown illumination position and pose. In this under-constrained setting, our approach is able to significantly outperform the baseline methods as much as 29\%. In this setting, the most difficult object to reconstruct is a chair, which often has thin structures in the shadow.  

Discriminative regression models are limited to produce reconstructions that are consistent with their training conditions, which is a principle restriction of prior methods. As we relax the number of known variables, the size of the input space significantly increases, which requires the regression baseline to increasingly generalize outside of the training set.  Table \ref{tab:unknown}  shows that regression is only marginally better than a nearest neighbor search on average. However, since our approach is generative, and integrates inductive biases about scene illumination, it is able to better generalize to more unconstrained settings. 

%We evaluate the regression model for generalizing to novel camera views and light source location. Specifically, during training, the input shadow images are taken with a fixed camera and light source location, thus creating a ``canonical' shadow. During inference, an input shadow taken with randomly sampled camera and light source location is used as input to the regression model. 
\begin{figure}[t]
    \centering
    \includegraphics[width=\linewidth]{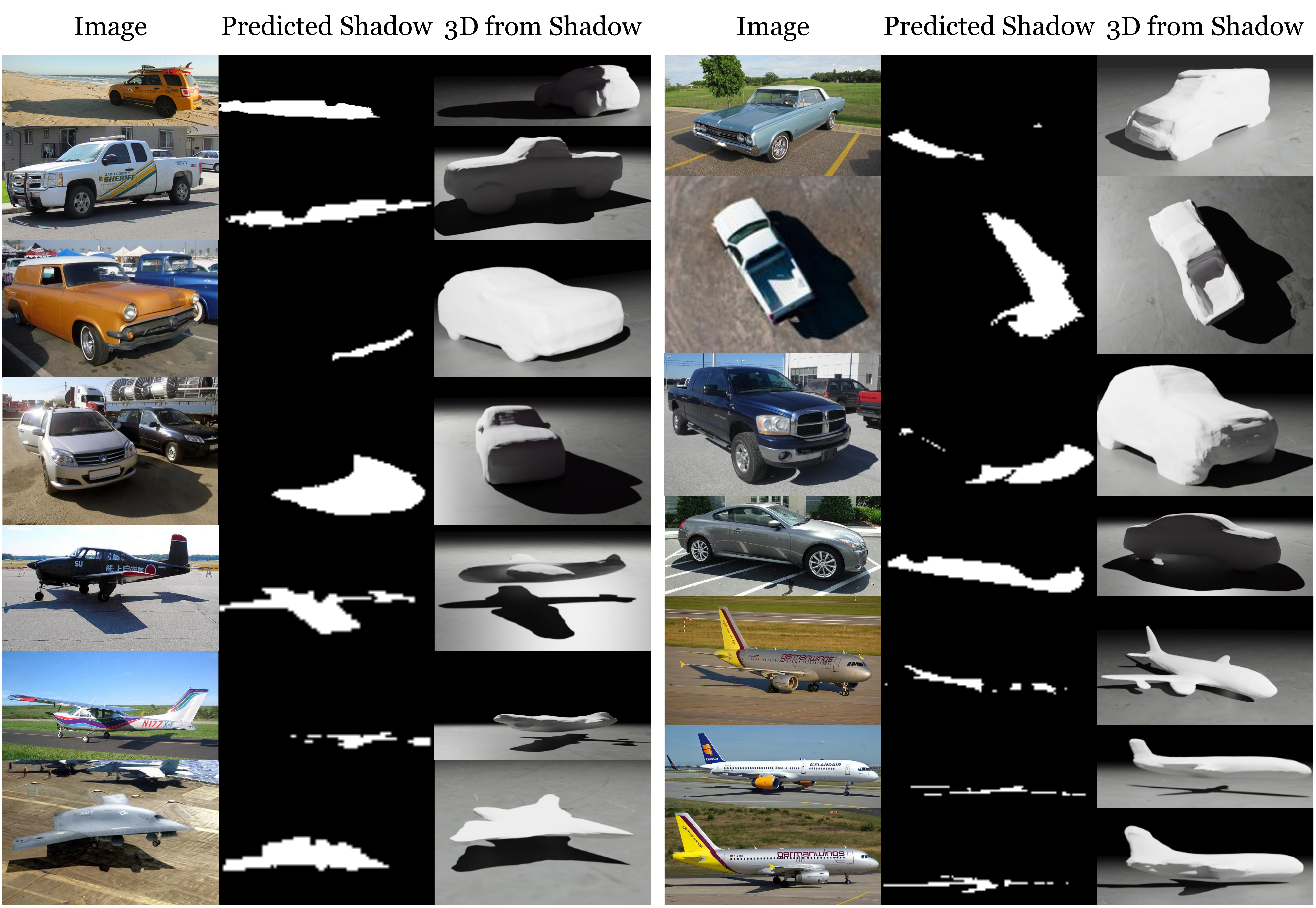}
    %\vspace{-2em}
    \caption{
    Qualitative results of 3D reconstructions in natural images. We first automatically segment shadow masks with~\cite{wang_instance_2020}. We then run our algorithm. 
    }
    \label{fig:soba}
\end{figure}
\subsubsection{Natural Image}
We applied our method to the real-world dataset in~\cite{wang_instance_2020}, and automatically obtain shadow segmentations with the detector proposed by the same work. Fig.~\ref{fig:soba} shows our 3D reconstructions from just the estimated shadows. Our method remains robust both for real-world images and slightly inaccurate shadow masks. 
These results also show our method estimates reasonable reconstructions when the ground-truth camera pose or light source location are unknown. Our method also returns reasonable-looking models even if the floor is not flat (e.g.\ car on sand).

\subsection{Diversity of Reconstructions}
\begin{figure}[t]
    \centering
    \includegraphics[width=\linewidth]{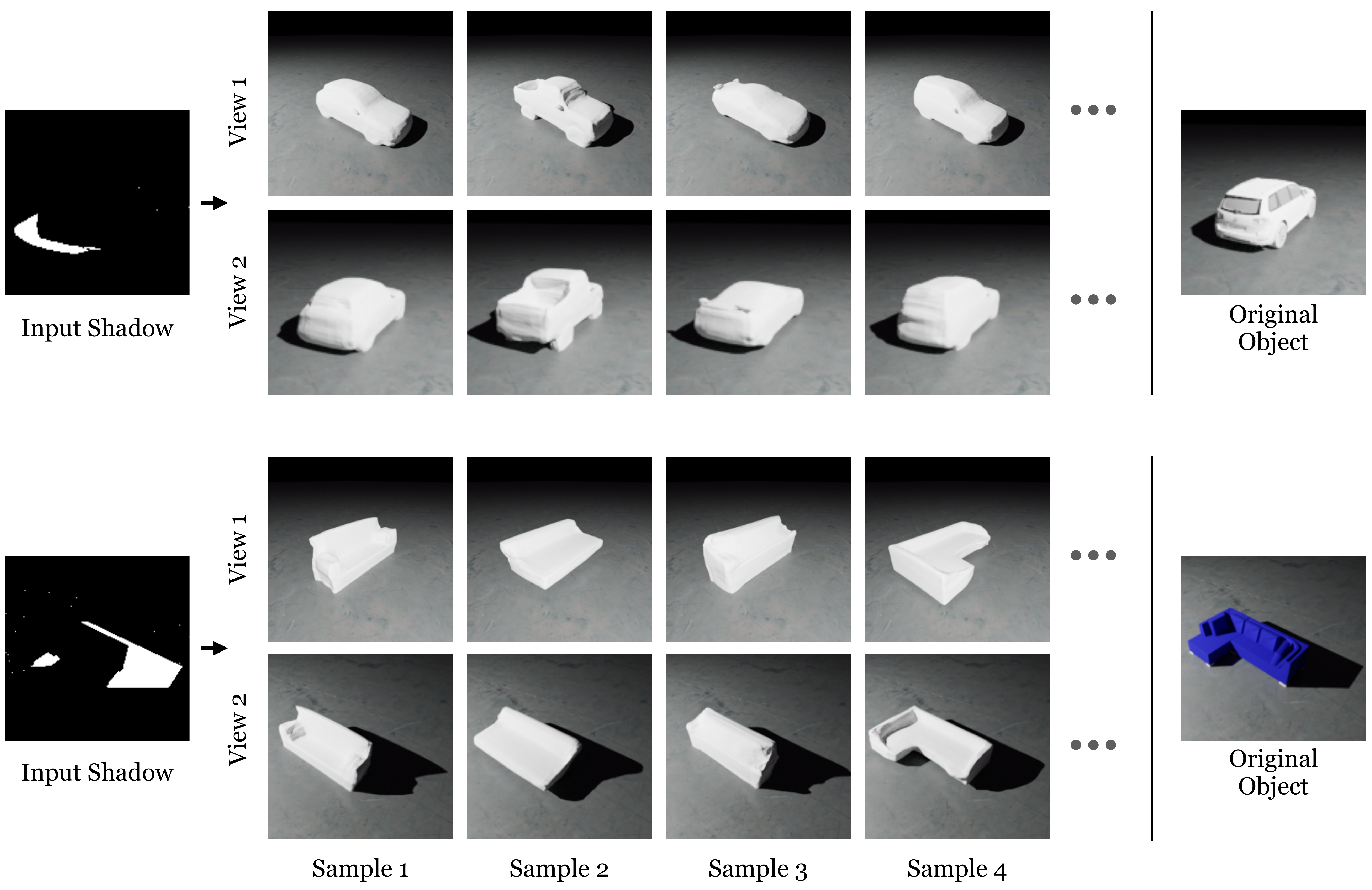}
    %\vspace{-2em}
    \caption{\textbf{Diversity of Reconstructions.} Given one shadow (\textbf{left}), our method is able to estimate multiple possible reconstructions (\textbf{middle}) that are consistent with the shadow. We show four samples from the model (columns), each under two different camera views (rows). The \textbf{right} side shows the original object.}
    \label{fig:diversity}
\end{figure}
\begin{SCfigure}[2][h]
    \centering
    \includegraphics[width=0.47\linewidth]{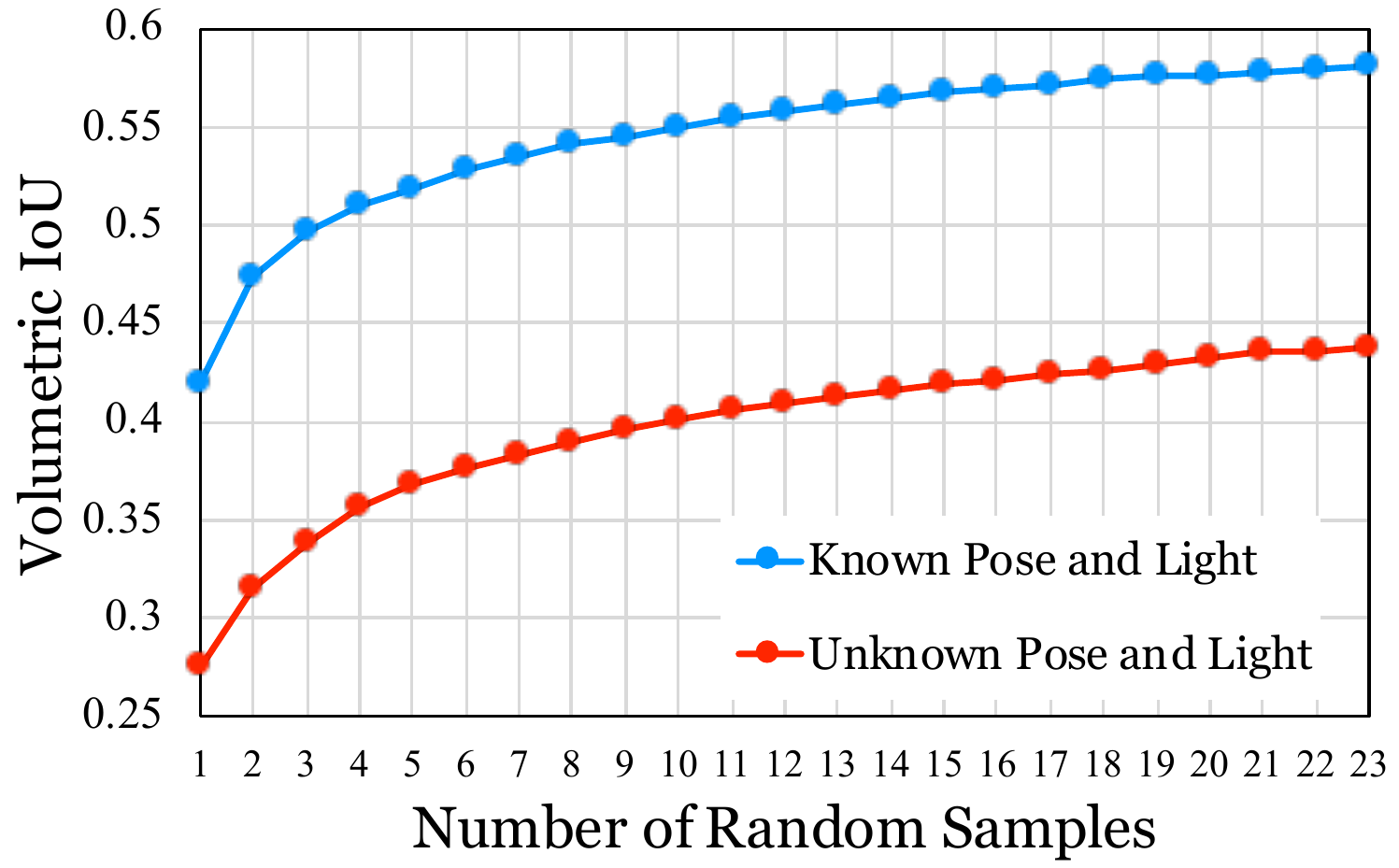}
    \caption{\textbf{Performance with Diverse Samples.} We show that our approach is able to make diverse 3D reconstructions from shadows. We plot best volumetric IoU versus the number of random samples from our method. The upward trends indicate the diversity of the prediction results from our method.}
    \label{fig:random}
\end{SCfigure}

By modeling the generative process of shadows, our approach is able to find multiple possible 3D shapes to explain the observed shadow.  When we sample different latent vectors as initialization, coupled with stochasticity from Gaussian noise in gradient descent, our method can generate a diverse set of solutions to minimize the shadow reconstruction loss.

Estimation of multiple possible scenes is an important advantage of our approach when compared with a regression model. There are many correct solutions to the 3D reconstruction task. When a regression model is trained to make a prediction for these tasks, the optimal solution is to predict the average of all the possible shapes in order to minimize the loss. In comparison, our approach does not regress to the mean under uncertainty.

%Instead of asking the model to predict a 3D shapes that minimizes reconstruction loss for all samples in the training dataset, we ask our model to just predict one 3D shape that's consistent with the shadow. In other words, our method is able to recover the specificity of the 3D shape prediction.\\

Figure \ref{fig:diversity} shows how the generative model is able to produce multiple, diverse samples that are all consistent with the shadow. For example, when given a shadow of a car, the method is able to produce both trucks and sedans that might cast the same shadow. When given the shadow of a sofa, the latent search discovers both L-shaped sofas and rectangular sofas that are compatible with the shadow.  Figure \ref{fig:random} quantitatively studies the diversity of samples from our method. As we perform latent search on the generative model with different random seeds, the likelihood of producing a correct prediction monotonically increases. This is valuable for our approach to be deployed in practice to resolve occlusions, such as robotics, where systems need to reason over all possible hypotheses for up-stream decision making.

\subsection{Analysis}
\begin{figure}[t!]
    \centering
    \includegraphics[width=\linewidth]{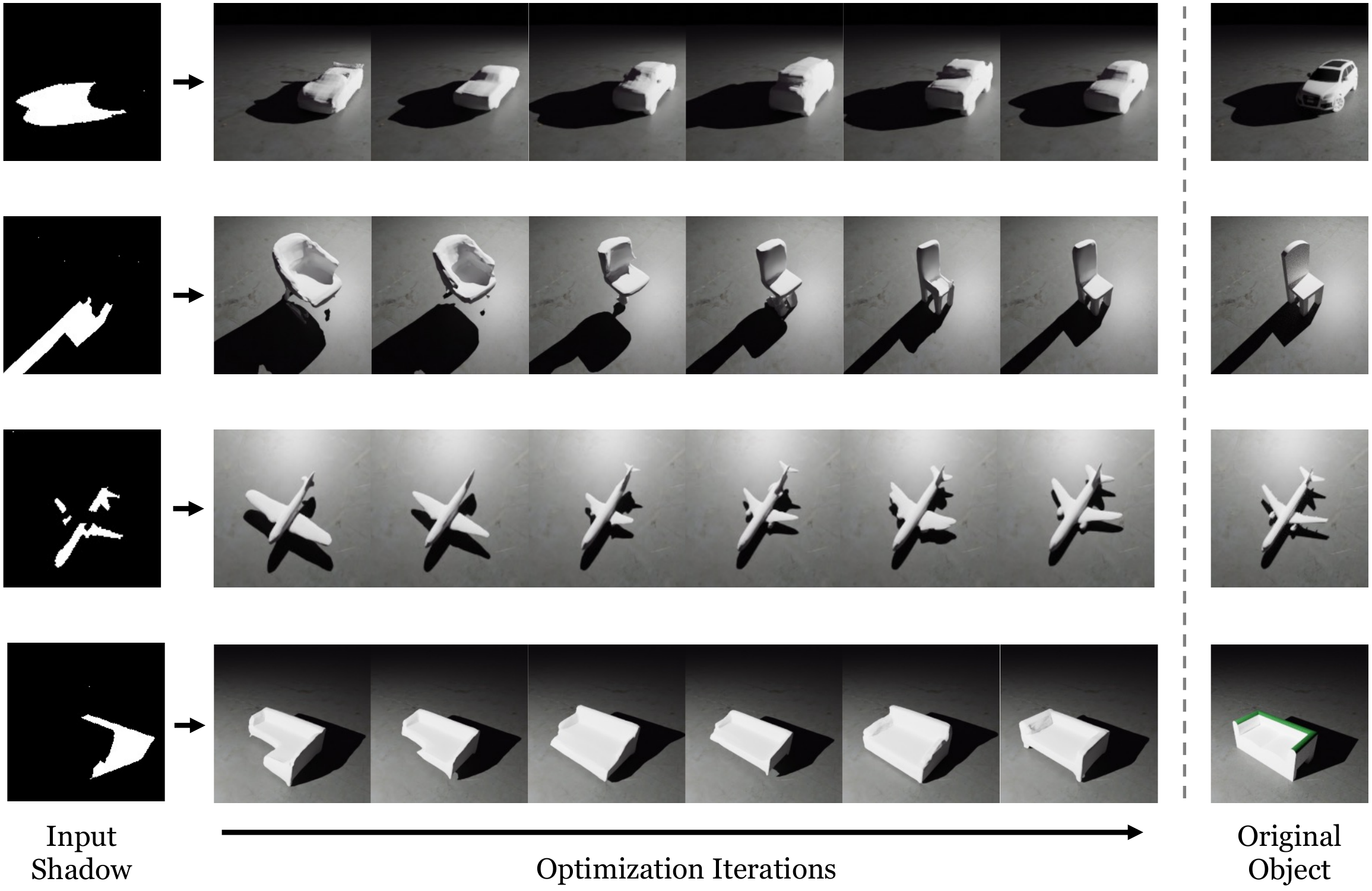}
    \vspace{-1em}
    \caption{\textbf{Visualizing Optimization Iterations.}  Visualizing the process of our model searching for 3D shapes that cast a shadow consistent with the input. The \textbf{1st row} shows the shadow used as a constraint for searching. \textbf{The middle} sequence of figure shows the process of searching in the latent space. The \textbf{last row} shows the original object as a reference, which is unseen by our model.\vspace{-1em}}
    \label{fig:optimization}
\end{figure}

\begin{figure}[t!]
    \centering
    \includegraphics[width=\linewidth]{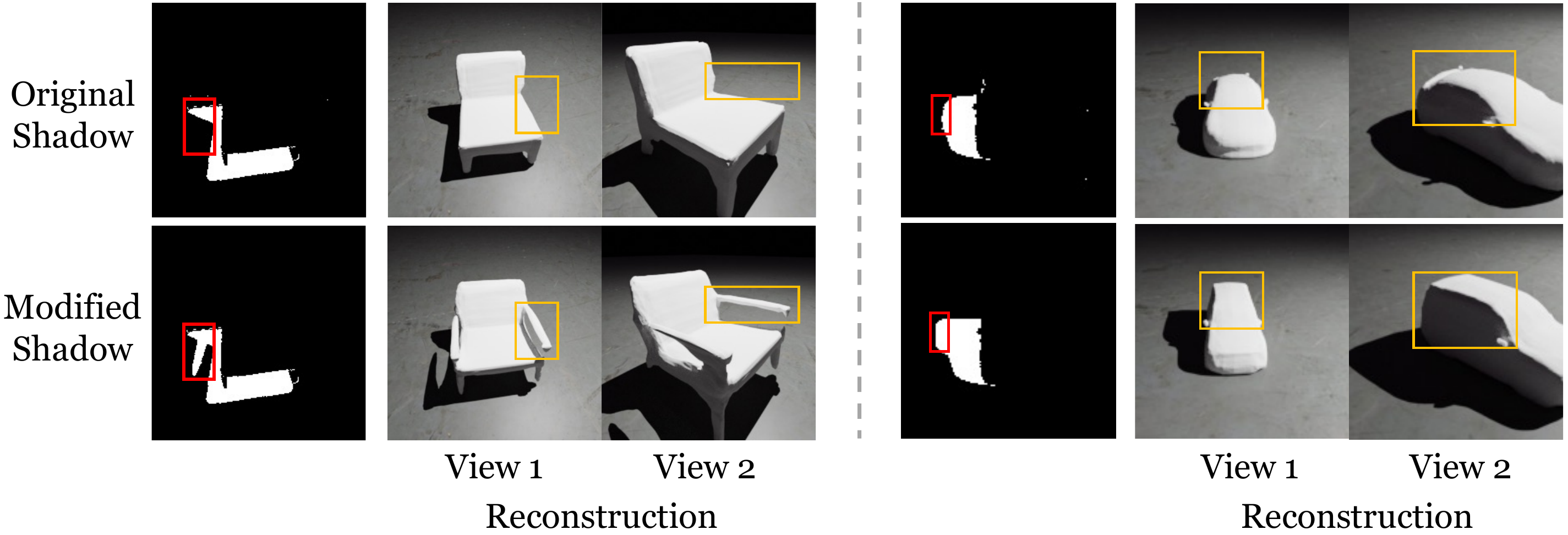}
    \vspace{-2em}
    \caption{\textbf{Reconstructing Manipulated Shadows.} We manually modify a shadow mask and comparing the reconstructed 3D object between the original and modified shadows. View 1 is the same as the original shadow image. View 2 is a second view for visualizing more details.}
    \label{fig:manipulation}
\end{figure}

\begin{SCfigure}[2][t!]
    \centering
    \includegraphics[width=0.55\linewidth]{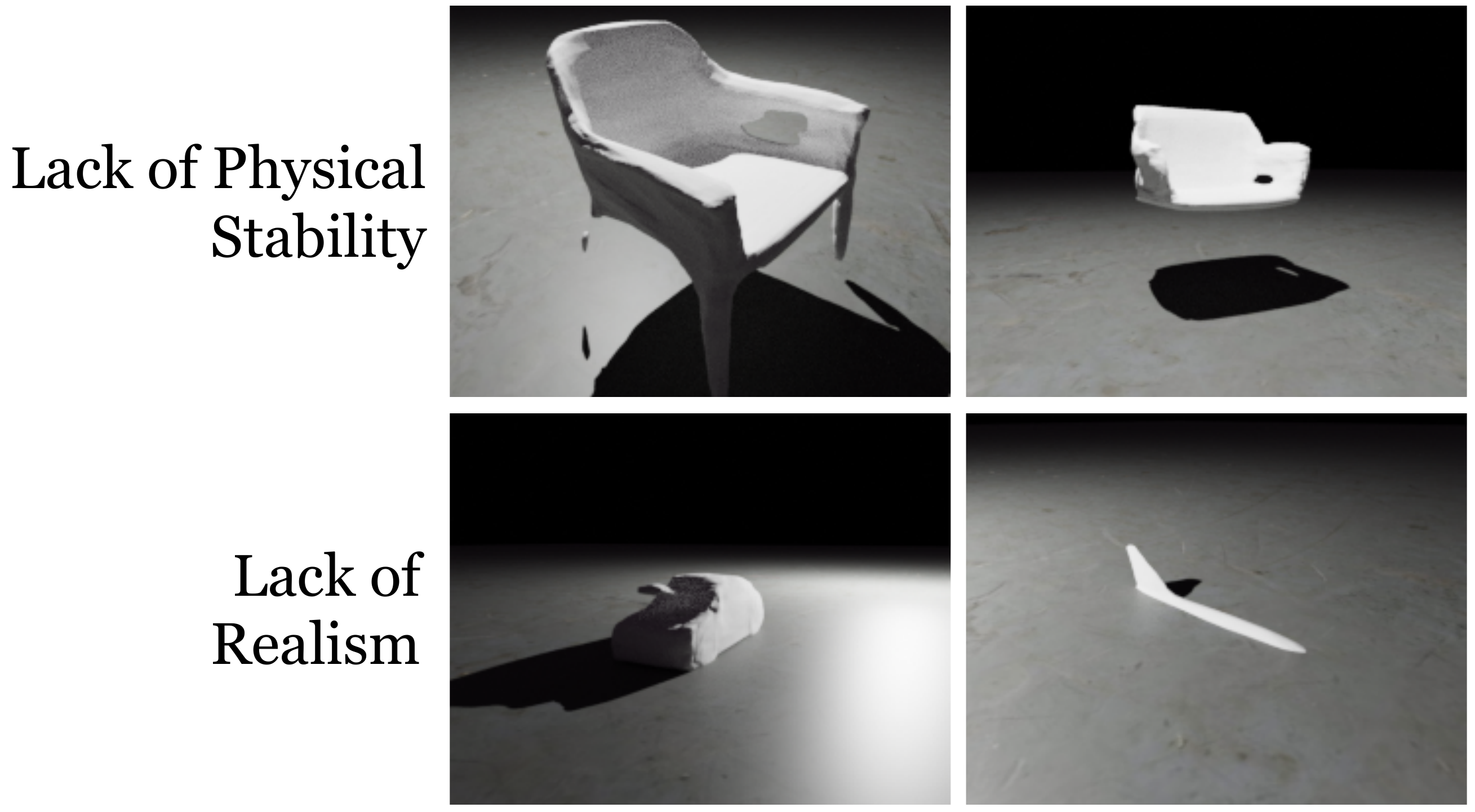}
    \caption{\textbf{Failures.} We visualize representative failures where the model produces incorrect shapes that still match the shadow. Our experiments suggest that results can be further improved with more priors, such as physical knowledge (\textbf{top}) and refined generative models (\textbf{bottom}).}
    \label{fig:failures}
\end{SCfigure}

\textbf{Optimization Process:}
To gain intuition into how our model progresses in the latent space to reach the final shadow-consistent reconstruction, we visualize in Figure \ref{fig:optimization} the optimization process by extracting the meshes corresponding to several optimization iterations before converging at the end. Figure \ref{fig:optimization} shows a clear transition from the first mesh generated from a randomly sampled latent vector, to the last mesh that accurately cast shadows matching the input ones. The reconstructed meshes at the end also match the original objects.

\textbf{Reconstructions of Modified Shadows:}
We found that our approach is able to exploit subtle details in shadows in order to produce accurate reconstructions. To study this behavior, we manually made small modifications to some of the shadow images, and analyzed how the resulting reconstructions changed. Figure \ref{fig:manipulation} shows two examples. In the example on the left, we modified the shadow of a chair to add an arm rest in the shadow image. In the comparison between the original reconstruction and modified reconstruction, we can see an arm rest being added to the reconstructed chair. In the example on the right, we take a shadow image of a sedan and make the shadow corresponding to the rear trunk part higher. The reconstructed car from the modified image becomes an SUV to adapt to the modified shadow.

\textbf{Analysis of Failures:} We show a few representative examples of failures in Figure \ref{fig:failures}. Although these shapes match the provided shadow, they are incorrect because they either lack physical stability or produce objects that are unlikely to be found in the natural visual world. These failures suggest that further progress on this problem is possible by integrating more extensive priors about objects and physics.

\section{Conclusions}

This paper shows that generative models are a promising mechanism to explain shadows in images. Our experiments show that jointly searching the latent space of a generative model and parameters for the light position and object pose allows us to reconstruct 3D objects from just an observation of the shadow. We believe tightly integrating empirical priors about objects with models of image formation will be an excellent avenue for resolving occlusions in scenes.\\

\textbf{Acknowledgements:} This research is based on work supported by Toyota Research Institute, the NSF CAREER Award \#2046910, and the DARPA MCS program under Federal Agreement No. N660011924032. SM is supported by the NSF GRFP fellowship. The views and conclusions contained herein are those of the authors and should not be interpreted as necessarily representing the official policies, either expressed or implied, of the sponsors.

%Quantitative results and qualitative visualizations show that our approach is able to generate 

%This paper shows that generative models of objects and differentiable rendering of shadows

%We presented a framework to invert the shadow generation process in order to reconstruct a 3D object from just its shadow by searching in the latent space of a 3D generative model for shadow-consistent shapes. Under very limited constraints, our model can generate a diverse set of possible 3D shapes that produce the same shadow. We further showcase that this method can be used to reconstruct objects that are partially or even completely occluded when the shadow of the object is visible. While we obtained exciting quantitative and qualitative results, our approach has several limitations. For example, given the binary nature of fa shadow mask, our model cannot reconstruct transparent or semi-transparent object such as glass due to the difficulty of shadow detection in these cases. Additionally, we can improve the quality of the reconstruction results by adding a discriminator trained adversarially. With these limitations and challenges, we believe our method remains a important step to use the shadows in a scene to infer the 3D structure of an object, and more generally, a framework to perform 3D reconstruction by searching in the latent space of a generative model under constraints.

%\clearpage
\bibliographystyle{splncs04}
\bibliography{egbib,references}
\end{document}

%% file: def.tex
\def\Blue{\color{blue}}
\def\Purple{\color{purple}}

\def\A{\bm{A}}
\def\a{{\bf a}}
\def\B{{\bf B}}
\def\b{{\bf b}}
\def\C{{\bf C}}
\def\c{{\bf c}}
\def\D{{\bf D}}
\def\d{{\bf d}}
\def\E{{\bf E}}
\def\e{{\bf e}}
\def\f{{\bf f}}
\def\F{{\bf F}}
\def\K{{\bf K}}
\def\k{{\bf k}}
\def\L{{\bf L}}
\def\H{{\bf H}}
\def\h{{\bf h}}
\def\G{{\bf G}}
\def\g{{\bf g}}
\def\I{{\bf I}}
\def\J{{\bf J}}
\def\R{{\bf R}}
\def\X{{\bf X}}
\def\Y{{\bf Y}}
\def\OO{{\bf O}}
\def\oo{{\bf o}}
\def\P{{\bf P}}
\def\p{{\bf p}}
\def\Q{{\bf Q}}
\def\q{{\bf q}}
\def\r{{\bf r}}
\def\s{{\bf s}}
\def\S{{\bf S}}
\def\t{{\bf t}}
\def\T{{\bf T}}
\def\x{{\bf x}}
\def\y{{\bf y}}
\def\z{{\bf z}}
\def\Z{{\bf Z}}
\def\M{{\bf M}}
\def\m{{\bf m}}
\def\n{{\bf n}}
\def\U{{\bf U}}
\def\u{{\bf u}}
\def\V{{\bf V}}
\def\v{{\bf v}}
\def\W{{\bf W}}
\def\w{{\bf w}}
\def\0{{\bf 0}}
\def\1{{\bf 1}}

\def\AM{{\mathcal A}}
\def\EM{{\mathcal E}}
\def\FM{{\mathcal F}}
\def\TM{{\mathcal T}}
\def\UM{{\mathcal U}}
\def\XM{{\mathcal X}}
\def\YM{{\mathcal Y}}
\def\NM{{\mathcal N}}
\def\OM{{\mathcal O}}
\def\IM{{\mathcal I}}
\def\GM{{\mathcal G}}
\def\PM{{\mathcal P}}
\def\LM{{\mathcal L}}
\def\MM{{\mathcal M}}
\def\DM{{\mathcal D}}
\def\SM{{\mathcal S}}
\def\ZM{{\mathcal Z}}
\def\RB{{\mathbb R}}
\def\EB{{\mathbb E}}
\def\VB{{\mathbb V}}

\def\tx{\tilde{\bf x}}
\def\ty{\tilde{\bf y}}
\def\tz{\tilde{\bf z}}
\def\hd{\hat{d}}
\def\HD{\hat{\bf D}}
\def\hx{\hat{\bf x}}
\def\hR{\hat{R}}

\def\alp{\mbox{\boldmath$\alpha$\unboldmath}}
\def\Ome{\mbox{\boldmath$\omega$\unboldmath}}
\def\Om{\mbox{\boldmath$\Omega$\unboldmath}}
\def\bet{\mbox{\boldmath$\beta$\unboldmath}}
\def\et{\mbox{\boldmath$\eta$\unboldmath}}
\def\ep{\mbox{\boldmath$\epsilon$\unboldmath}}
\def\ph{\mbox{\boldmath$\phi$\unboldmath}}
\def\Pii{\mbox{\boldmath$\Pi$\unboldmath}}
\def\pii{\mbox{\boldmath$\pi$\unboldmath}}
\def\Ph{\mbox{\boldmath$\Phi$\unboldmath}}
\def\Ps{\mbox{\boldmath$\Psi$\unboldmath}}
\def\tha{\mbox{\boldmath$\theta$\unboldmath}}
\def\Tha{\mbox{\boldmath$\Theta$\unboldmath}}
\def\muu{\mbox{\boldmath$\mu$\unboldmath}}
\def\Si{\mbox{\boldmath$\Sigma$\unboldmath}}
\def\si{\mbox{\boldmath$\sigma$\unboldmath}}
\def\Gam{\mbox{\boldmath$\Gamma$\unboldmath}}
\def\gamm{\mbox{\boldmath$\gamma$\unboldmath}}
\def\Lam{\mbox{\boldmath$\Lambda$\unboldmath}}
\def\De{\mbox{\boldmath$\Delta$\unboldmath}}
\def\vps{\mbox{\boldmath$\varepsilon$\unboldmath}}
\def\Up{\mbox{\boldmath$\Upsilon$\unboldmath}}
\def\xii{\mbox{\boldmath$\xi$\unboldmath}}
\def\Xii{\mbox{\boldmath$\Xi$\unboldmath}}
\def\Lap{\mbox{\boldmath$\LM$\unboldmath}}
\newcommand{\ti}[1]{\tilde{#1}}

\def\tr{\mathrm{tr}}
\def\etr{\mathrm{etr}}
\def\etal{{\em et al.\/}\,}
\newcommand{\indep}{{\;\bot\!\!\!\!\!\!\bot\;}}
\def\argmax{\mathop{\rm argmax}}
\def\argmin{\mathop{\rm argmin}}
\def\vec{\text{vec}}
\def\cov{\text{cov}}
\def\dg{\text{diag}}

% \newtheorem{observation}{\textbf{Observation}}
% \newtheorem{remark}{Remark}
% \newtheorem{theorem}{Theorem}
% \newtheorem{lemma}{Lemma}
% \newtheorem{definition}{Definition}
% \newtheorem{problem}{Problem}
% \newtheorem{proposition}{Proposition}
% \newtheorem{cor}{Corollary}
% \numberwithin{theorem}{section}
% \numberwithin{lemma}{section}
% \numberwithin{remark}{section}
% \numberwithin{cor}{section}
% \numberwithin{proposition}{section}

\newtheorem{assumption}{Assumption}

\newcommand{\tabref}[1]{Table~\ref{#1}}
\newcommand{\secref}[1]{Sec.~\ref{#1}}
\newcommand{\figref}[1]{Fig.~\ref{#1}}
\newcommand{\lemref}[1]{Lemma~\ref{#1}}
\newcommand{\thmref}[1]{Theorem~\ref{#1}}
\newcommand{\clmref}[1]{Claim~\ref{#1}}
\newcommand{\crlref}[1]{Corollary~\ref{#1}}
\newcommand{\asuref}[1]{Assumption~\ref{#1}}
\newcommand{\eqnref}[1]{Eqn.~\ref{#1}}
\newcommand{\algref}[1]{Alg.~\ref{#1}}

\renewcommand{\tilde}{\widetilde}
\renewcommand{\hat}{\widehat}
\renewcommand{\frac}{\tfrac}

%% file: eccv2022submission.bbl
\begin{thebibliography}{10}
\providecommand{\url}[1]{\texttt{#1}}
\providecommand{\urlprefix}{URL }
\providecommand{\doi}[1]{https://doi.org/#1}

\bibitem{agarwal2010bundle}
Agarwal, S., Snavely, N., Seitz, S.M., Szeliski, R.: Bundle adjustment in the
  large. In: European conference on computer vision. pp. 29--42. Springer
  (2010)

\bibitem{bleyer2011patchmatch}
Bleyer, M., Rhemann, C., Rother, C.: Patchmatch stereo-stereo matching with
  slanted support windows. In: Bmvc. vol.~11, pp. 1--11 (2011)

\bibitem{bouguet1993shadows}
Bouguet, J.Y., Perona, P.: 3d photography using shadows in dual-space geometry.
  International Journal of Computer Vision  \textbf{35}(2),  129--149 (1999)

\bibitem{broadhurst2001probabilistic}
Broadhurst, A., Drummond, T.W., Cipolla, R.: A probabilistic framework for
  space carving. In: Proceedings eighth IEEE international conference on
  computer vision. ICCV 2001. vol.~1, pp. 388--393. IEEE (2001)

\bibitem{brock_generative_2016}
Brock, A., Lim, T., Ritchie, J.M., Weston, N.: Generative and {Discriminative}
  {Voxel} {Modeling} with {Convolutional} {Neural} {Networks}. arXiv:1608.04236
  [cs, stat]  (Aug 2016), \url{http://arxiv.org/abs/1608.04236}, arXiv:
  1608.04236

\bibitem{chang2015shapenet}
Chang, A.X., Funkhouser, T., Guibas, L., Hanrahan, P., Huang, Q., Li, Z.,
  Savarese, S., Savva, M., Song, S., Su, H., et~al.: Shapenet: An
  information-rich 3d model repository. arXiv preprint arXiv:1512.03012  (2015)

\bibitem{de1999poxels}
De~Bonet, J.S., Viola, P.: Poxels: Probabilistic voxelized volume
  reconstruction. In: Proceedings of International Conference on Computer
  Vision (ICCV). pp. 418--425. Citeseer (1999)

\bibitem{dong_image_2015}
Dong, C., Loy, C.C., He, K., Tang, X.: Image {Super}-{Resolution} {Using}
  {Deep} {Convolutional} {Networks}. arXiv:1501.00092 [cs]  (Jul 2015),
  \url{http://arxiv.org/abs/1501.00092}, arXiv: 1501.00092

\bibitem{fan_point_2016}
Fan, H., Su, H., Guibas, L.: A {Point} {Set} {Generation} {Network} for {3D}
  {Object} {Reconstruction} from a {Single} {Image}. arXiv:1612.00603 [cs]
  (Dec 2016), \url{http://arxiv.org/abs/1612.00603}, arXiv: 1612.00603

\bibitem{galliani2016gipuma}
Galliani, S., Lasinger, K., Schindler, K.: Gipuma: Massively parallel
  multi-view stereo reconstruction. Publikationen der Deutschen Gesellschaft
  f{\"u}r Photogrammetrie, Fernerkundung und Geoinformation e. V
  \textbf{25}(361-369), ~2 (2016)

\bibitem{ucmrGoel20}
Goel, S., Kanazawa, A., , Malik, J.: Shape and viewpoints without keypoints.
  In: ECCV (2020)

\bibitem{groueix2018papier}
Groueix, T., Fisher, M., Kim, V.G., Russell, B.C., Aubry, M.: A
  papier-m{\^a}ch{\'e} approach to learning 3d surface generation. In:
  Proceedings of the IEEE conference on computer vision and pattern
  recognition. pp. 216--224 (2018)

\bibitem{hartley2003multiple}
Hartley, R., Zisserman, A.: Multiple view geometry in computer vision.
  Cambridge university press (2003)

\bibitem{hoiem2005automatic}
Hoiem, D., Efros, A.A., Hebert, M.: Automatic photo pop-up. In: ACM SIGGRAPH
  2005 Papers, pp. 577--584 (2005)

\bibitem{horry1997tour}
Horry, Y., Anjyo, K.I., Arai, K.: Tour into the picture: using a spidery mesh
  interface to make animation from a single image. In: Proceedings of the 24th
  annual conference on Computer graphics and interactive techniques. pp.
  225--232 (1997)

\bibitem{h36m_pami}
Ionescu, C., Papava, D., Olaru, V., Sminchisescu, C.: Human3.6m: Large scale
  datasets and predictive methods for 3d human sensing in natural environments.
  IEEE Transactions on Pattern Analysis and Machine Intelligence
  \textbf{36}(7),  1325--1339 (jul 2014)

\bibitem{cmrKanazawa18}
Kanazawa, A., Tulsiani, S., Efros, A.A., Malik, J.: Learning category-specific
  mesh reconstruction from image collections. In: ECCV (2018)

\bibitem{humanMotionKanazawa19}
Kanazawa, A., Zhang, J.Y., Felsen, P., Malik, J.: Learning 3d human dynamics
  from video. In: Computer Vision and Pattern Regognition (CVPR) (2019)

\bibitem{kim_accurate_2016}
Kim, J., Lee, J.K., Lee, K.M.: Accurate {Image} {Super}-{Resolution} {Using}
  {Very} {Deep} {Convolutional} {Networks}. In: 2016 {IEEE} {Conference} on
  {Computer} {Vision} and {Pattern} {Recognition} ({CVPR}). pp. 1646--1654.
  IEEE, Las Vegas, NV, USA (Jun 2016). \doi{10.1109/CVPR.2016.182},
  \url{http://ieeexplore.ieee.org/document/7780551/}

\bibitem{krull2015learning}
Krull, A., Brachmann, E., Michel, F., Yang, M.Y., Gumhold, S., Rother, C.:
  Learning analysis-by-synthesis for 6d pose estimation in rgb-d images. In:
  Proceedings of the IEEE international conference on computer vision. pp.
  954--962 (2015)

\bibitem{li2020self}
Li, X., Liu, S., Kim, K., Mello, S.D., Jampani, V., Yang, M.H., Kautz, J.:
  Self-supervised single-view 3d reconstruction via semantic consistency. In:
  European Conference on Computer Vision. pp. 677--693. Springer (2020)

\bibitem{liu2019soft}
Liu, S., Li, T., Chen, W., Li, H.: Soft rasterizer: A differentiable renderer
  for image-based 3d reasoning. In: Proceedings of the IEEE/CVF International
  Conference on Computer Vision. pp. 7708--7717 (2019)

\bibitem{menon2020pulse}
Menon, S., Damian, A., Hu, S., Ravi, N., Rudin, C.: Pulse: Self-supervised
  photo upsampling via latent space exploration of generative models. In:
  Proceedings of the ieee/cvf conference on computer vision and pattern
  recognition. pp. 2437--2445 (2020)

\bibitem{mescheder2019occupancy}
Mescheder, L., Oechsle, M., Niemeyer, M., Nowozin, S., Geiger, A.: Occupancy
  networks: Learning 3d reconstruction in function space. In: Proceedings of
  the IEEE/CVF Conference on Computer Vision and Pattern Recognition. pp.
  4460--4470 (2019)

\bibitem{mildenhall2020nerf}
Mildenhall, B., Srinivasan, P.P., Tancik, M., Barron, J.T., Ramamoorthi, R.,
  Ng, R.: Nerf: Representing scenes as neural radiance fields for view
  synthesis. In: European conference on computer vision. pp. 405--421. Springer
  (2020)

\bibitem{niemeyer2020differentiable}
Niemeyer, M., Mescheder, L., Oechsle, M., Geiger, A.: Differentiable volumetric
  rendering: Learning implicit 3d representations without 3d supervision. In:
  Proceedings of the IEEE/CVF Conference on Computer Vision and Pattern
  Recognition. pp. 3504--3515 (2020)

\bibitem{ongie_deep_2020}
Ongie, G., Jalal, A., Metzler, C.A., Baraniuk, R.G., Dimakis, A.G., Willett,
  R.: Deep {Learning} {Techniques} for {Inverse} {Problems} in {Imaging}.
  arXiv:2005.06001 [cs, eess, stat]  (May 2020),
  \url{http://arxiv.org/abs/2005.06001}, arXiv: 2005.06001

\bibitem{pontes2018image2mesh}
Pontes, J.K., Kong, C., Sridharan, S., Lucey, S., Eriksson, A., Fookes, C.:
  Image2mesh: A learning framework for single image 3d reconstruction. In:
  Asian Conference on Computer Vision. pp. 365--381. Springer (2018)

\bibitem{sadekar2022shadow}
Sadekar, K., Tiwari, A., Raman, S.: Shadow art revisited: a differentiable
  rendering based approach. In: Proceedings of the IEEE/CVF Winter Conference
  on Applications of Computer Vision. pp. 29--37 (2022)

\bibitem{savarese_3d_2007}
Savarese, S., Andreetto, M., Rushmeier, H., Bernardini, F., Perona, P.: {3D}
  {Reconstruction} by {Shadow} {Carving}: {Theory} and {Practical}
  {Evaluation}. International Journal of Computer Vision  \textbf{71}(3),
  305--336 (Mar 2007). \doi{10.1007/s11263-006-8323-9},
  \url{https://doi.org/10.1007/s11263-006-8323-9}

\bibitem{schonberger2016pixelwise}
Sch{\"o}nberger, J.L., Zheng, E., Frahm, J.M., Pollefeys, M.: Pixelwise view
  selection for unstructured multi-view stereo. In: European Conference on
  Computer Vision. pp. 501--518. Springer (2016)

\bibitem{seitz2006comparison}
Seitz, S.M., Curless, B., Diebel, J., Scharstein, D., Szeliski, R.: A
  comparison and evaluation of multi-view stereo reconstruction algorithms. In:
  2006 IEEE computer society conference on computer vision and pattern
  recognition (CVPR'06). vol.~1, pp. 519--528. IEEE (2006)

\bibitem{seitz1999photorealistic}
Seitz, S.M., Dyer, C.R.: Photorealistic scene reconstruction by voxel coloring.
  International Journal of Computer Vision  \textbf{35}(2),  151--173 (1999)

\bibitem{shafer_using_1983}
Shafer, S.A., Kanade, T.: Using shadows in finding surface orientations.
  Computer Vision, Graphics, and Image Processing  \textbf{22}(1),  145--176
  (Apr 1983). \doi{10.1016/0734-189X(83)90099-3},
  \url{https://www.sciencedirect.com/science/article/pii/0734189X83900993}

\bibitem{shin_pixels_2018}
Shin, D., Fowlkes, C.C., Hoiem, D.: Pixels, {Voxels}, and {Views}: {A} {Study}
  of {Shape} {Representations} for {Single} {View} {3D} {Object} {Shape}
  {Prediction}. In: 2018 {IEEE}/{CVF} {Conference} on {Computer} {Vision} and
  {Pattern} {Recognition}. pp. 3061--3069. IEEE, Salt Lake City, UT (Jun 2018).
  \doi{10.1109/CVPR.2018.00323},
  \url{https://ieeexplore.ieee.org/document/8578421/}

\bibitem{troccoli_shadow_2004}
Troccoli, A., Allen, P.: A {Shadow} {Based} {Method} for {Image} to {Model}
  {Registration}. In: 2004 {Conference} on {Computer} {Vision} and {Pattern}
  {Recognition} {Workshop}. pp. 169--169 (Jun 2004).
  \doi{10.1109/CVPR.2004.289}

\bibitem{waltz_understanding_1975}
Waltz, D.: Understanding {Line} {Drawings} of {Scenes} with {Shadows}. In: The
  {Psychology} of {Computer} {Vision}. p. pages. McGraw-Hill (1975)

\bibitem{wang_instance_2020}
Wang, T., Hu, X., Wang, Q., Heng, P.A., Fu, C.W.: Instance {Shadow}
  {Detection}. In: 2020 {IEEE}/{CVF} {Conference} on {Computer} {Vision} and
  {Pattern} {Recognition} ({CVPR}). pp. 1877--1886. IEEE, Seattle, WA, USA (Jun
  2020). \doi{10.1109/CVPR42600.2020.00195},
  \url{https://ieeexplore.ieee.org/document/9157490/}

\bibitem{sgld}
Welling, M., Teh, Y.W.: Bayesian learning via stochastic gradient langevin
  dynamics. In: Proceedings of the 28th international conference on machine
  learning (ICML-11). pp. 681--688. Citeseer (2011)

\bibitem{wu2020unsupervised}
Wu, S., Rupprecht, C., Vedaldi, A.: Unsupervised learning of probably symmetric
  deformable 3d objects from images in the wild. In: Proceedings of the
  IEEE/CVF Conference on Computer Vision and Pattern Recognition. pp. 1--10
  (2020)

\bibitem{ye2021shelf}
Ye, Y., Tulsiani, S., Gupta, A.: Shelf-supervised mesh prediction in the wild.
  In: Proceedings of the IEEE/CVF Conference on Computer Vision and Pattern
  Recognition. pp. 8843--8852 (2021)

\bibitem{yu2020pixelnerf}
Yu, A., Ye, V., Tancik, M., Kanazawa, A.: pixelnerf: Neural radiance fields
  from one or few images. In: CVPR (2021)

\bibitem{yuille2006vision}
Yuille, A., Kersten, D.: Vision as bayesian inference: analysis by synthesis?
  Trends in cognitive sciences  \textbf{10}(7),  301--308 (2006)

\end{thebibliography}
